\documentclass[final]{cvpr}

\usepackage{times}
\usepackage{epsfig}
\usepackage{graphicx}
\usepackage{amsmath}
\usepackage{amssymb}
\usepackage{url}            
\usepackage{booktabs}       
\usepackage{amsfonts}       
\usepackage{pifont}
\usepackage[dvipsnames,table,xcdraw]{xcolor}
\usepackage{pifont}
\usepackage{multirow}
\usepackage{overpic}
\usepackage{listings}

\usepackage{float}
\usepackage{algorithm} 
\usepackage{algpseudocode}

\newcommand{\nameofmethod}{VOLO}
\newcommand{\nameoflayer}{Outlooker}

\definecolor{mygreen}{RGB}{0,100,0}
\definecolor{myblue}{RGB}{10,100,200}
\definecolor{myred}{RGB}{200,0,0}

\DeclareMathOperator{\outlookatt}{OutlookAtt}
\DeclareMathOperator{\LN}{LN}
\DeclareMathOperator{\MLP}{MLP}
\DeclareMathOperator{\softmax}{Softmax}
\DeclareMathOperator{\matmul}{MatMul}

\newcommand{\highlight}[1]{\textcolor{ForestGreen}{\textbf{#1}}}
\newcommand{\myPara}[1]{\vspace{.05in}\noindent\textbf{#1:}}

\usepackage[pagebackref=false,colorlinks=true,linkcolor=myred,citecolor=RoyalBlue,bookmarks=false]{hyperref}

\def \pzo {\phantom{0}} 
\def \dzo {\phantom{00}}

\def\cvprPaperID{***} 
\def\confYear{CVPR 2021}

\title{VOLO:  Vision Outlooker for Visual Recognition}

\author{Li Yuan$^{1,2}$\thanks{Equal contribution.} \qquad Qibin Hou$^2$\footnotemark[1] \qquad Zihang Jiang$^2$ \qquad Jiashi Feng$^{1,2}$ \qquad Shuicheng Yan$^1$ \\
  $^1$Sea AI Lab \qquad $^2$National University of Singapore \\
  {\tt\small \{ylustcnus,andrewhoux,jzh0103\}@gmail.com, \{fengjs, yansc\}@sea.com} \\
}

\begin{document}

\maketitle

\begin{abstract}

Visual recognition has been dominated by convolutional neural networks (CNNs) for years.
Though recently the prevailing vision transformers (ViTs) have shown great potential of self-attention based models in ImageNet classification, their performance is still inferior to that of  the latest SOTA CNNs if no extra data are provided.
In this work, we try to close the performance gap and demonstrate that attention-based models are indeed able to outperform CNNs.
We find a major factor limiting the performance of ViTs for ImageNet classification is their low efficacy in encoding fine-level features into the token representations.
To resolve this, we introduce a novel \emph{outlook attention} and present a   simple and general architecture, termed Vision Outlooker (VOLO).
Unlike self-attention that focuses on global dependency modeling at a coarse level,
the {outlook attention} efficiently encodes finer-level features and contexts into tokens, which is shown to be critically beneficial to recognition performance but largely ignored by the self-attention.
Experiments show that our VOLO achieves 87.1\% top-1 accuracy on ImageNet-1K classification, which is the first model exceeding 87\% accuracy on this competitive benchmark, without using any extra training data.
In addition, the pre-trained  \nameofmethod{} transfers well to downstream tasks, such as semantic segmentation.
We achieve 84.3\% mIoU score on the cityscapes validation set and
54.3\% on the ADE20K validation set.
Code is available at \url{https://github.com/sail-sg/volo}.

\end{abstract}

\section{Introduction}

Modeling in visual recognition, which was long dominated by convolutional neural networks (CNNs), has recently been revolutionized by Vision Transformers (ViTs)~\cite{dosovitskiy2020image, touvron2020training, yuan2021tokens}. 
Different from CNNs that aggregate and transform features via local and dense convolutional kernels, ViTs directly model long-range dependencies of local patches (\emph{a.k.a.}~tokens) through the self-attention mechanism which is with greater flexibility in modeling visual contents. 
Despite the remarkable effectiveness on visual recognition~\cite{liu2021swin,jiang2021token,touvron2021going,zhou2021refiner}, the performance of ViT models still lags behind that of the state-of-the-art CNN models.
For instance, as shown in Table~\ref{tab:comp_sota}, the state-of-the-art transformer-based CaiT~\cite{touvron2021going} attains 86.5\% top-1 accuracy on ImageNet, which however is still 0.3\% lower compared with the 86.8\% top-1 accuracy achieved by the CNN-based NFNet-F5~\cite{brock2021high} with SAM and augmult~\cite{foret2020sharpness,fort2021drawing}.

\begin{figure}[t]
    \centering
    \small
    \includegraphics[width=\linewidth]{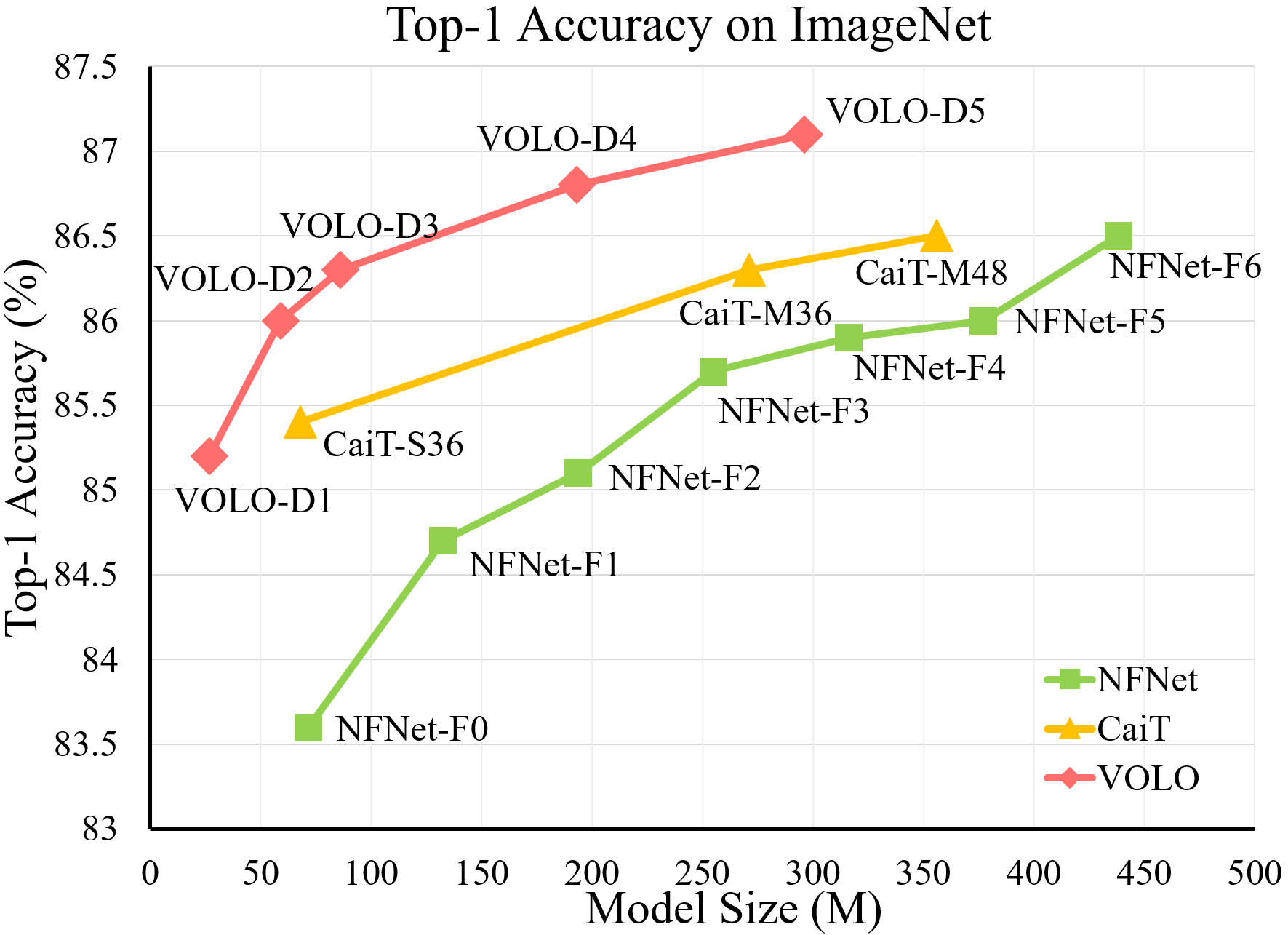}
    \caption{ImageNet top-1 accuracy of state-of-the-art CNN-based and 
    Transformer-based models. All the results are obtained based on the best test resolutions, without using any extra training data. Our \nameofmethod{}-D5 achieves the best accuracy, outperforming the latest NFNet-F6 w/ SAM \cite{brock2021high,foret2020sharpness} and CaiT-M48
    w/ KD \cite{hinton2015distilling,yuan2020revisiting}, while using much less training parameters. 
    To our best knowledge, \nameofmethod{}-D5 is the first model exceeding 87\% top-1 accuracy on ImageNet.}
    \label{fig:teaser}
\end{figure}

\begin{table*}[tp!]
  \centering
  \small
  \setlength\tabcolsep{1.5mm}
  \renewcommand\arraystretch{1}
  \caption{Comparison with previous state-of-the-art classification models,
  most of which have once achieved leading positions on the leaderboard of PaperWithCode\protect \footnotemark (w/o extra data).}
  \label{tab:comp_sota}
  \begin{tabular}{lcccccc} \toprule
    Settings & LV-ViT \cite{jiang2021token} & CaiT~\cite{touvron2021going} & NFNet-F6 \cite{brock2021high} & NFNet-F5 \cite{brock2021high} & \nameofmethod{}-D5 (Ours) \\ \toprule
    Test Resolution & $448 \times 448$ & $448 \times 448$ & $576 \times 576$ & $544 \times 544$ & $448 \times 448$ / $512 \times 512$ \\
    Model Size & 140M & 356M & 438M & 377M & 296M \\
    Computations & 157B & 330B & 377B & 290B & 304B / 412B \\
    Architecture & Vision Transformer & Vision Transformer & Convolutions & Convolutions & \nameofmethod{} \\
    Extra Augmentations & Token Labeling~\cite{jiang2021token} & Knowledge Distill & SAM \cite{foret2020sharpness} & SAM + augmult \cite{foret2020sharpness,fort2021drawing} & Token Labeling~\cite{jiang2021token} \\ \midrule[0.5pt]
    ImageNet Top-1 Acc. & 86.4 & 86.5 & 86.5 & 86.8 & \textbf{87.0} / \textbf{87.1} \\
    \bottomrule
  \end{tabular}
\end{table*}

In this work we try to close such performance gap. 
We find one major factor limiting ViTs from outperforming CNNs is their low efficacy in encoding fine-level features and contexts into token representations, which are critical for achieving compelling visual recognition performance.
Fine-level information can be encoded into tokens by finer-grained image tokenization, which however would lead to a token sequence of greater length that increases quadratically the complexity of the self-attention mechanism of ViTs. 

In this work, we present a new simple and light-weight attention mechanism, termed
\emph{\nameoflayer{}}, to enrich the token representations with fine level information efficiently.
The proposed \nameoflayer{} innovates the way of generating attention for token aggregation, and enables the model to efficiently encode fine-level information.
In particular, it extrapolates the mechanism of aggregating surrounding tokens from the anchor token feature directly via efficient linear projections, thus getting rid of the expensive dot-product attention computation.

Based on the proposed \nameoflayer{}, we present \nameofmethod{}, a simple yet powerful model architecture for visual recognition.
\nameofmethod{} achieves fine-level token representation encoding and global information aggregation with a two-stage  architecture design.
Specifically, given an input image of size $224\times224$, before using self-attention to build global dependencies at the coarse level (\eg, $14\times14$), the \nameofmethod{}   tokenizes the image on smaller-size  patches (\eg, $8\times8$) and employs multiple \nameoflayer{}s to encode token representations at the fine level (\eg, $28\times28$).
The obtained token representations are more expressive, thus significantly improving the model performance in image classification.

\footnotetext{{\url{https://paperswithcode.com/sota/image-classification-on-imagenet}}}

Experiments show that our proposed \nameofmethod{} performs extremely well in ImageNet classification.
Take a \nameofmethod{} model with 26.6M learnable parameters as an example.
It achieves 84.2\% top-1 accuracy on ImageNet without using any extra data.
Finetuning this model on the $384\times384$ input resolution can further increase the accuracy to 85.2\%.
Moreover, when scaling up the model size to 296M parameters, it can reach a top-1 accuracy of 87.1\% on ImageNet, 90.6\% on ImageNet-ReaL, and 78.0\% on ImageNet-V2, setting new SOTA performance for all the three classification benchmarks.

As depicted in Figure~\ref{fig:teaser}, compared to the previous state-of-the-art CNN-based model (NFNet-F6 \cite{brock2021high} with SAM \cite{foret2020sharpness}), and the transformer-based model (CaiT-M48 \cite{touvron2021going} with KD), our best model \nameofmethod{}-D5 leverages the least amount of learnable parameters but achieves the best accuracy.
Moreover, as shown in Table~\ref{tab:comp_sota}, even compared with previous state-of-the-art models using stronger data augmentation and optimization methods (such as SAM~\cite{foret2020sharpness}
and augmult \cite{fort2021drawing}), our \nameoflayer{} still performs the best.

Our \nameofmethod{} also achieves strong performance on the semantic segmentation task.
We run experiments on two widely-used segmentation benchmarks: Cityscapes~\cite{cordts2016cityscapes} and ADE20K~\cite{zhou2019semantic}.
Experiments show that our \nameofmethod{} attains 84.3\% mIoU score on the Cityscapes validation set, 0.3\% better than the previous state-of-the-art result (by SegFormer-B5~\cite{xie2021segformer}).
On the ADE20K validation set, we achieve 54.3\% mIoU score, largely improving the state-of-the-art result (53.5\%) by Swin Transformer~\cite{liu2021swin}, which is pretrained on ImageNet-22k.

\section{Method}

Our model can be regarded as an architecture with two separate stages. 
The first stage consists of a stack of \nameoflayer{}s that generates fine-level token representations.
The second stage deploys a sequence of transformer blocks  to aggregate global information.
At the beginning of each stage, a patch embedding module is used to map the input to token representations with designed shapes.

\subsection{\nameoflayer{}}

\nameoflayer{} consists of an outlook attention layer for spatial information encoding and a multi-layer perceptron (MLP) for inter-channel information interaction.
Given a sequence of input $C$-dim  token representations $\mathbf{X} \in \mathbb{R}^{H \times W \times C}$, \nameoflayer{} can be written as follows:
\begin{align}
    &\tilde{\mathbf{X}} = \outlookatt(\LN(\mathbf{X})) + \mathbf{X}, \\
    &\mathbf{Z} = \MLP(\LN(\tilde{\mathbf{X}})) + \tilde{\mathbf{X}}.
\end{align}
Here, $\LN$ refers to LayerNorm \cite{liu2020rethinking}.

\subsubsection{Outlook Attention}

\begin{figure*}[t]
    \centering
    \small
    \includegraphics[width=0.9\linewidth]{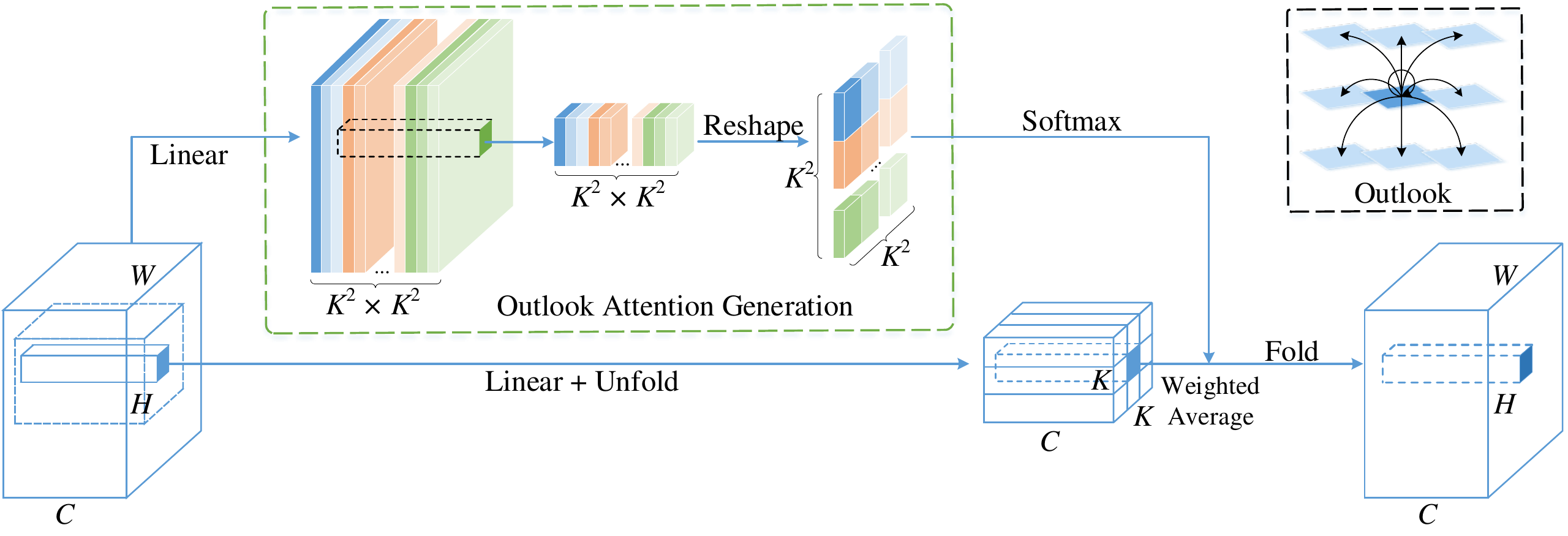}
    \caption{Illustration of outlook attention. 
    The outlook attention matrix for a local window of size $K \times K$  can be simply generated from the center token with a linear layer followed by a reshape operation  (highlighted by the green dash box).
    As the attention weights are generated from the center token within the window and act on the neighbor tokens and itself (as demonstrated in the black dash block), we name these operations as outlook attention.}
    \label{fig:outlook}
\end{figure*}

\begin{algorithm}[t]
\caption{{Outlook attention code (PyTorch-like)}}
\label{alg:outloot_att}
\definecolor{codeblue}{rgb}{0.25,0.5,0.25}
\lstset{
	backgroundcolor=\color{white},
	basicstyle=\fontsize{7.2pt}{7.2pt}\ttfamily\selectfont,
	columns=fullflexible,
	breaklines=true,
	captionpos=b,
	commentstyle=\fontsize{7.2pt}{7.2pt}\color{codeblue},
	keywordstyle=\fontsize{7.2pt}{7.2pt},
}
\begin{lstlisting}[language=python]
# H: height, W: width, K: kernel size
# x: input tensor (H, W, C)

def init()
    v_pj = nn.Linear(C, C)
    attn = nn.Linear(C, k ** 4)
    unfold = nn.Unfold(K, padding)
    fold = nn.Fold(output_size=(H, W), K, padding)

def outlook_attention(x):   # code in forward
    v = v_pj(x).permute(2, 1, 0)
    
    # Eqn. (3), embedding set of neighbors
    v = unfold(v).reshape(C, K*K, H*W).permute(2, 1, 0)
    a = attn(x).reshape(H*W, K*K, K*K)
    
    # Eqn. (4), weighted average
    a = a.softmax(dim=-1)
    x = mul(a, v).permute(2, 1, 0).reshape(C*K*K, H*W)
    
    # Eqn. (5)
    x = fold(x).permute(2, 1, 0)

    return x
\end{lstlisting}
\end{algorithm}

Outlook attention is simple, efficient, and easy to implement. The main insights behind it are: 
1) the  feature at each spatial location is representative enough to generate attention weights for locally aggregating its neighboring features; 
2) the dense and local  spatial aggregation can encode fine-level information efficiently.  

For each spatial location $(i, j)$, outlook attention computes
its similarity to all the neighbors within a local window of size $K \times K$
centered at $(i, j)$.
Unlike self-attention that requires a Query-Key matrix multiplication for the computation of the attention 
(i.e., $\softmax(\mathbf{Q}^\top \mathbf{K} /\sqrt{d})$),
outlook attention simplifies this process via just a reshaping operation.

Formally, given the input $\mathbf{X}$, each $C$-dim token  is first projected, using two linear layers of weights $\mathbf{W}_A\in \mathbb{R}^{C \times K^4}$
and $\mathbf{W}_V\in \mathbb{R}^{C \times C}$, into   outlook weights
$\mathbf{A}\in \mathbb{R}^{H \times W \times K^4}$ and value representation $\mathbf{V}\in \mathbb{R}^{H \times W \times C}$, 
respectively.
Let $\mathbf{V}_{\Delta_{i,j}} \in \mathbb{R}^{C \times K^2}$ denote all the values within the local window
centered at $(i, j)$, \ie,
\begin{equation} \label{eqn:unfold}
\mathbf{V}_{\Delta_{i,j}}=\{\mathbf{V}_{i+p-\lfloor \frac{K}{2} \rfloor,j+q-\lfloor \frac{K}{2} \rfloor}\},  \quad  0 \leq p,q <K.
\end{equation}
\textbf{Outlook attention} The outlook weight at location $(i,j)$ is directly used as the attention weight for value aggregation, by reshaping it
to $\hat{\mathbf{A}}_{i,j} \in \mathbb{R}^{K^2\times K^2}$, followed by a $\softmax$ function.
Thus, the value projection procedure can be written as
\begin{equation}
    \mathbf{Y}_{\Delta_{i,j}} = \matmul(\softmax(\hat{\mathbf{A}}_{i,j}), \mathbf{V}_{\Delta_{i,j}}).
\end{equation}
\textbf{Dense aggregation} Outlook attention aggregates the projected value representations densely. Summing up the different weighted values at the same location from different local windows yields the output
\begin{equation} \label{eqn:fold}
    \tilde{\mathbf{Y}}_{i,j} = \sum_{0 \leq m, n < K} \mathbf{Y}_{\Delta_{i+m-\lfloor \frac{K}{2}\rfloor,j+n-\lfloor \frac{K}{2}\rfloor}}^{i,j}.
\end{equation}
PyTorch-like outlook attention codes are summarized in Algorithm~\ref{alg:outloot_att}.
Eqn.~\eqref{eqn:unfold} and Eqn.~\eqref{eqn:fold} correspond to the \texttt{Unfold} and \texttt{Fold}
operations, respectively.
After outlook attention, a linear layer is often adopted as in self-attention.

\subsubsection{Multi-Head Outlook Attention}

The implementation of multi-head outlook attention is simple.
Suppose the head number is set to $N$.
We just need to adjust the weight shape of $\mathbf{W}_A$ such that  $\mathbf{W}_A\in \mathbb{R}^{C \times N \cdot K^4}$.
Then, the outlook weight and value embeddings are uniformly split into $N$ segments, yielding
$\mathbf{A}_n \in \mathbb{R}^{H \times W \times K^4}$ and $\mathbf{V}_n\in \mathbb{R}^{H \times W \times C_N}, \{n=1,2,...,N\}$, 
where $C_N$ is the dimension of each head which satisfies $C_N \times N = C$.
For each $(\mathbf{A}_n, \mathbf{V}_n)$ pair, the outlook attention is separately computed, which are then concatenated as the output of the multi-head outlook attention.
In our experiment section, we will ablate the impact of the head number on model performance.

\begin{table*}[h]
  \centering
  \small
  \setlength\tabcolsep{1.6mm}
  \renewcommand\arraystretch{1.0}
  \caption{Architecture information of different variants of \nameofmethod{}. The resolution information is based
  on an input image of size $224\times224$. 
  The number of parameters includes that of weights for both the network backbone and the classifier head. 
  `Layer' refers to either a \nameoflayer{} block or a Transformer block.}
  \label{tab:arch_details}
  \begin{tabular}{lcccccc} \toprule
    Specification & \nameofmethod{}-D1 & \nameofmethod{}-D2 & \nameofmethod{}-D3 & \nameofmethod{}-D4 & \nameofmethod{}-D5 \\ \toprule
    Patch Embedding & $8 \times 8$ & $8 \times 8$ & $8 \times 8$ & $8 \times 8$ & $8 \times 8$ \\ \midrule
    Stage 1 ($28 \times 28$) & $\begin{bmatrix} \text{head: 6, stride: 2} \\ \text{kernel: } 3\times3 \\ \text{mlp: 3, dim: 192}  \end{bmatrix} $ & $\begin{bmatrix} \text{head: 8, stride: 2} \\ \text{kernel: } 3\times3 \\ \text{mlp: 3, dim: 256} \end{bmatrix} $ & $\begin{bmatrix} \text{head: 8, stride: 2} \\ \text{kernel: } 3\times3 \\ \text{mlp: 3, dim: 256} \end{bmatrix} $ & $\begin{bmatrix} \text{head: 12, stride: 2} \\ \text{kernel: } 3\times3 \\ \text{mlp: 3, dim: 384} \end{bmatrix} $ & $\begin{bmatrix} \text{head: 12, stride: 2} \\ \text{kernel: } 3\times3 \\ \text{mlp: 4, dim: 384} \end{bmatrix} $ \\ 
     & $\times 4$ & $\times 6$ & $\times 8$ & $\times 8$ & $\times 12$\\ \toprule
    Patch Embedding & $2 \times 2$ & $2 \times 2$ & $2 \times 2$ & $2 \times 2$ & $2 \times 2$ \\ \midrule
    Stage 2 ($14 \times 14$) & $\begin{bmatrix} \text{\#heads: 12} \\ \text{mlp: 3, dim: 384} \end{bmatrix} $ & $\begin{bmatrix} \text{\#heads: 16} \\ \text{mlp: 3, dim: 512} \end{bmatrix} $ &  $\begin{bmatrix} \text{\#heads: 16} \\ \text{mlp: 3, dim: 512} \end{bmatrix} $ & $\begin{bmatrix} \text{\#heads: 16} \\ \text{mlp: 3, dim: 768} \end{bmatrix}  $ & $\begin{bmatrix} \text{\#heads: 16} \\ \text{mlp: 4, dim: 768} \end{bmatrix} $ \\ 
     & $\times 14$ & $\times 18$ & $\times 28$ & $\times 28$ & $\times 36$ \\ \midrule
    Total Layers & 18 & 24 & 36 & 36 & 48 \\
    Parameters & 26.6M & 58.7M & 86.3M & 193M & 296M \\ 
    \bottomrule
  \end{tabular}
\end{table*}

\subsubsection{Discussion}

Our outlook attention inherits the merits of both convolutions and self-attention.
It offers the following advantages.
First of all, outlook attention encodes spatial information by measuring the similarity between pairs of token representations, which is more parameter-efficient for feature learning than convolutions, 
as studied in previous work~\cite{liu2021swin,srinivas2021bottleneck}.
Second, outlook attention adopts a sliding window mechanism to locally encode token representations at fine level, and to some extent  preserves the crucial positional information  for vision tasks \cite{hou2021coordinate,wang2020axial}. 
Third, the way of generating attention weights is simple and efficient.
Unlike self-attention that relies on a query-key matrix multiplication, our outlook weight can be directly produced by a simple reshaping operation, saving computation.
To see this, we compare the computation for a self-attention (SA) and that for a local version of self-attention (LSA)
when operating on $H\times W$ tokens with a sliding window size $K \times K$:
\begin{align}
    \text{M-Adds}(\textbf{SA}) &\approx 4HWC^2 + 2(HW)^2C \\
    \text{M-Adds}(\textbf{LSA}) &\approx 4HWC^2 + 2HWK^2C \\
    \text{M-Adds}(\textbf{OA}) &\approx HWC(2C + NK^4) + HWK^2C.
\end{align}
Considering a normal case in which $C=384$, $K=3$, and $N=6$,
our outlook attention is more computationally efficient as $NK^4 < 2C$.

\subsection{Network Architecture Variants}

We build the proposed \nameofmethod{} based on the LV-ViT model \cite{jiang2021token}
which we find is a surprisingly strong baseline that achieves 86.2\% ImageNet top-1 accuracy with 150M learnable parameters.
The original LV-ViT model consists of a patch embedding module that maps an input image of size $224\times224$ to $14\times14$ tokens and a sequence of transformers that operate on the $14\times14$ tokens.
To leverage the fine-level token representations, in the first stage, we adjust the patch embedding module to make the image tokenize on small image patches of size $8\times8$ instead of $16\times16$.
A stack of \nameoflayer{}s is used to generate more expressive token representations at the fine level.
In the second stage, another patch embedding module is utilized to downsample the tokens.
A sequence of transformers is then adopted to encode global information.

Based on the above network structure, we introduce five versions of the proposed \nameofmethod{}:
\nameofmethod{}-D1, \nameofmethod{}-D2, \nameofmethod{}-D3, \nameofmethod{}-D4, and \nameofmethod{}-D5.
Detailed hyper-parameter settings of all the five versions can be found in Table~\ref{tab:arch_details}.
In all versions, we keep the ratio of Outlooker and Transformer to around 1:3, which we have empirically found works the best in our experiments. We also add two class attention layers~\cite{touvron2021going} in the final stage to update the class embedding.
The hidden dimension in Outlookers is set to half of that in Transformers.

\section{Experiments}

We evaluate our proposed \nameofmethod{} on the ImageNet \cite{deng2009imagenet} dataset.
During training, we do not use any extra training data.
Our code is based on \texttt{PyTorch}~\cite{paszke2019pytorch}, the \texttt{Token Labeling toolbox}~\cite{jiang2021token}, and \texttt{timm}~\cite{rw2019timm}.
We use the LV-ViT-S~\cite{jiang2021token} model with Token Labeling as our baseline.

\begin{table}[t]
  \centering
  \small
  \setlength\tabcolsep{2mm}
  \renewcommand\arraystretch{1}
  \caption{Model settings. We use a linear learning rate scaling strategy
  $lr=\text{LR}_{\text{base}}\cdot \frac{\text{batch\_size}}{\text{1024}}$. For all models, we
  set the batch size to 1024. }
  \label{tab:settings}
  \begin{tabular}{lcccccc} \toprule
    Specification & D1 & D2 & D3 & D4 & D5 \\ \midrule[0.5pt] 
    MLP Ratio & 3 & 3 & 3 & 3 & 4 \\
    Parameters & 27M & 59M & 86M & 193M & 296M \\
    Stoch. Dep. Rate &0.1&0.2&0.5&0.5&0.75 \\
    Crop Ratio &0.96&0.96&0.96&1.15&1.15 \\
    LR$_{\text{base}}$ & 1.6e-3  & 1e-3 & 1e-3 & 1e-3 & 8e-4  \\
    weight decay & 5e-2  & 5e-2 & 5e-2 & 5e-2 &5e-2 \\
    \bottomrule
  \end{tabular}
\end{table}

\begin{table*}[htp!]
    \centering
    \setlength\tabcolsep{1.9mm}
    \renewcommand\arraystretch{1}
    \caption{Top-1 accuracy comparison of our method with previous state-of-the-art methods on ImageNet \cite{deng2009imagenet},
    ImageNet Real~\cite{beyer2020we}, and ImageNet-V2~\cite{recht2019imagenet}. We split the results into
    5 segments according the model size. All models are trained without external data. 
    With the same computation and parameter constraint, our model consistently outperforms
    other MLP-like, CNN-based, and transformer-based counterparts. `Train size' and `Test size' refer to
    resolutions used in training and finetuning (test for CNNs). Our \nameofmethod{}-D5 sets a new
    record on all three benchmarks, which is the first model attaining 87.1\% top-1 accuracy on ImageNet.}
    \label{tab:sota}
    \def \mysp {\hspace{7pt}}
    {\small 
    \begin{tabular}{l@{\ }@{\ }c|cc|cc|ccc}
    \toprule
    Network & Architecture & Params & FLOPs & Train size & Test size  &  Top-1  & Real Top-1 & V2 Top-1 \\
    \toprule

    DeiT-S~\cite{touvron2020training} & Transformer & \pzo22M & \dzo4.6B  & $224$ & $224$  & 79.9  & 85.7 & 68.5 \\
    T2T-ViT-14~\cite{yuan2021tokens} & Transformer  & \pzo22M & \dzo5.2B  & $224$ & $224$        & 81.5 &86.8  & 69.9 \\
    T2T-ViT-14$\uparrow$384~\cite{yuan2021tokens} & Transformer & \pzo22M & \pzo 17.1B  & $224$ & $384$ & 83.3 & 87.8 & 72.4 \\
    DeepViT-S~\cite{zhou2021deepvit} & Transformer & \pzo27M & \dzo6.2B & $224$ & $224$       & 82.3 & \_  & \_ \\
    ViP-Small/7~\cite{hou2021vision} & MLP-like & \pzo25M & \_ & 224 & 224 & 81.5 & \_ & \_ \\
    BoTNet-S1-59~\cite{srinivas2021bottleneck} & Hybrid & \pzo34M & \dzo7.3B & 224 & 224 & 81.7 & \_ & \_ \\
    EfficientNet-B5~\cite{tan2019efficientnet} & CNN  & \pzo30M & \dzo9.9B  & $456$ & $456$  & 83.6 & 88.3 & 73.6 \\
    LV-ViT-S$\uparrow$384~\cite{jiang2021token} & Transformer & \pzo26M & \pzo22.2B &  $224$ &  384  &  84.4 & 88.9 & 74.5 \\
     \nameofmethod{}-D1 & \nameofmethod{} & \pzo27M & \dzo6.8B &  \text{224} &  \text{224}  &  \text{84.2} & \text{89.0} & \text{74.0}\\
    \textbf{\nameofmethod{}-D1$\uparrow$384} & \textbf{\nameofmethod{}} & \textbf{\pzo27M} & \textbf{\pzo22.8B} &  \textbf{224} &  \textbf{384}  &  \textbf{85.2} & \textbf{89.6} & \textbf{75.6} \\
    \toprule
    
    CrossViT~\cite{chen2021crossvit} & Transformer & \pzo45M & \pzo56.6B & $224$ & $480$ & 84.1 & \_ & \_\\
    TNT-B~\cite{han2021transformer}  & Transformer & \pzo66M & \pzo14.1B & $224$ & $224$ & 82.8 & \_ & \_ \\
    ViP-Medium/7~\cite{hou2021vision} & MLP-like & \pzo55M & \_ & 224 & 224 & 82.7 & \_ & \_ \\
    DeepViT-L~\cite{zhou2021deepvit} & Transformer & \pzo55M & \pzo12.5B & $224$ & $224$ & 83.1 & \_ & \_ \\ 
    EfficientNet-B7~\cite{tan2019efficientnet}  & CNN  & \pzo66M & \pzo37.0B & $600$ & $600$  & 84.3 & \_ & \_ \\
    NFNet-F0~\cite{brock2021high}  & CNN & \pzo72M & \pzo12.4B & $192$ & $256$ & 83.6 & 88.1 & 72.6 \\
    CaiT-S36$\uparrow$384~\cite{touvron2021going} & Transformer & \pzo68M & \pzo48.0B & $224$ & $384$ &  85.4 & 89.8 & 76.2 \\
    LV-ViT-M$\uparrow$384~\cite{jiang2021token} & Transformer & \pzo56M & \pzo42.2B &  $224$ &  $384$ &  85.4 & 89.5 & 76.0 \\
    \text{\nameofmethod{}-D2} & \text{\nameofmethod{}} & \text{\pzo59M} & \text{\pzo14.1B} &  \text{224} &  \text{224}  &  \text{85.2} & \text{89.3} & \text{75.2}\\
    \textbf{\nameofmethod{}-D2$\uparrow$384} & \textbf{\nameofmethod{}} & \textbf{\pzo59M} & \textbf{\pzo46.1B} &  \textbf{224} &  \textbf{384}  &  \textbf{86.0} & \textbf{89.7}  & \textbf{76.4}\\
    \toprule
    
    ViT-B/16~\cite{dosovitskiy2020image}   & Transformer  & \pzo86M & \pzo55.4B & $224$ & $384$ & 77.9 & 83.6 & \_ \\
    DeiT-B~\cite{touvron2020training}      & Transformer  & \pzo86M & \pzo17.5B & $224$ & $224$ &  81.8 &   86.7 & \_ \\
    ViP-Large/7~\cite{hou2021vision} & MLP-like & \pzo88M & \_ & 224 & 224 & 83.2 & \_ & \_ \\
    Swin-B~\cite{liu2021swin}    & Transformer         & \pzo88M & \pzo47.0B   & $224$ & $384$        & 84.2 & \_ & \_ \\
    BoTNet-S1-128$\uparrow$384~\cite{srinivas2021bottleneck} & Hybrid & \pzo79M & \pzo45.8B & 256 & 384 & 84.7 & \_ & \_ \\
    Fix-EfficientNet-B8~\cite{tan2019efficientnet, touvron2019fixing} & CNN & \pzo87M & \pzo89.5B & $672$ & $800$  & 85.7 & 90.0 & \_ \\
    Refined-ViT-L$\uparrow$448~\cite{zhou2021refiner} & Transformer & \pzo81M & \pzo98.0B & 224 & 448 & 85.9 & \_ & \_ \\
    \text{\nameofmethod{}-D3} & \text{\nameofmethod{}} & \text{\pzo86M} & \text{\pzo20.6B} &  \text{224} &  \text{224}  &  \text{85.4} & \text{89.6} & \text{75.6}\\
    \textbf{\nameofmethod{}-D3$\uparrow$448} & \textbf{\nameofmethod{}} & \textbf{\pzo86M} & \textbf{\pzo67.9B} &  \textbf{224} &  \textbf{448}  &  \textbf{86.3} & \textbf{90.0} & \textbf{77.7} \\
    \toprule

    NFNet-F1~\cite{brock2021high} & CNN          & 133M    & \pzo35.5B & $224$ & $320$  & 84.7 & 88.9 & 74.4 \\
    NFNet-F2~\cite{brock2021high} & CNN          & 194M    & \pzo62.6B & $256$ & $352$  & 85.1 & 88.9 & 74.3 \\
    NFNet-F3~\cite{brock2021high}   & CNN        & 255M    & 115.0B    & $320$ & $416$  & 85.7 & 89.4 & 75.2 \\
    \text{\nameofmethod{}-D4} & \text{\nameofmethod{}} & \text{193M} & \text{\pzo43.8B} &  \text{224} &  \text{224}  &  \text{85.7} & \text{89.7} & \text{75.6}\\
    \textbf{\nameofmethod{}-D4$\uparrow$448} & \textbf{\nameofmethod{}} & \textbf{193M} & \textbf{\pzo197B} &  \textbf{224} &  \textbf{448}  &  \textbf{86.8} & \textbf{90.5} & \textbf{77.8}\\
    \toprule
    
    NFNet-F4~\cite{brock2021high}   & CNN        & 316M    & 215B    & $384$ & $512$   & 85.9 & 89.4 & 75.2 \\
    NFNet-F5~\cite{brock2021high}   & CNN        & 377M    & 290B    & $416$ & $544$   & 86.0 & 89.2 & 74.6 \\
    NFNet-F6~\cite{brock2021high}+SAM   & CNN        & 438M    & 377B    & $448$ & $576$   & 86.5 & 89.2 & 75.8 \\
    ViT-L/16~\cite{dosovitskiy2020image} & Transformer & 307M    & 191B    & $224$ & $384$  & 76.5 & 82.2 & \_ \\
    CaiT-M36$\uparrow$448~\cite{touvron2021going} & Transformer & 271M & 248B & $224$ & $448$  & 86.3 & 90.2 & 76.7\\
    CaiT-M48$\uparrow$448~\cite{touvron2021going} & Transformer & 356M & 330B & $224$ & $448$  & 86.5 & 90.2 & 76.9\\
    \text{\nameofmethod-D5} & \text{\nameofmethod{}} & \text{296M} & \text{69.0B} &  \text{224} &  \text{224}  &  \text{86.1} & \text{89.9} & \text{76.3} \\
    \textbf{\nameofmethod-D5$\uparrow$448} & \textbf{\nameofmethod{}} & \textbf{296M} & \textbf{304B} &  \textbf{224} &  \textbf{448}  &  \textbf{87.0} & \textbf{90.6} & \textbf{77.8} \\
    \textbf{\nameofmethod-D5$\uparrow$512} & \textbf{\nameofmethod{}} & \textbf{296M} & \textbf{412B} &  \textbf{224} &  \textbf{512}  &  \textbf{87.1} & \textbf{90.6} & \textbf{78.0} \\
    \bottomrule
    \end{tabular}}
\end{table*}

\myPara{Setup}
We use the AdamW optimizer~\cite{loshchilov2017decoupled} with a linear learning rate scaling strategy 
$lr = \text{LR}_{\text{base}} \times \frac{\text{batch}\_\text{size}}{1024}$ and $5\times 10^{-2}$ weight decay rate as suggested by
previous work \cite{touvron2020training,jiang2021token}, and $\text{LR}_{\text{base}}$ are given in Table~\ref{tab:settings} for all VOLO models.
Stochastic Depth~\cite{huang2016deep} is used.
We train our models on the ImageNet dataset for 300 epochs.
For data augmentation methods, we use CutOut~\cite{zhong2020random}, RandAug~\cite{cubuk2020randaugment},
and the Token Labeling objective with MixToken \cite{jiang2021token}.
We do not use MixUp~\cite{zhang2017mixup} or CutMix~\cite{yun2019cutmix} as they conflict with MixToken.
We train all \nameofmethod{} models on a machine node with 8 NVIDIA V100 or A100 GPUs except for \nameofmethod{}-D5 which needs two nodes. For \nameofmethod{}-D1 and \nameofmethod{}-D2, 4 GPUs also suffice with batch size 512 (16G) or 1024 (32G).
For finetuning on larger image resolutions, we set the batch size to 512, learning rate to 5e-6, weight decay 
to 1e-8 and run the models for 30 epochs.
Other hyper-parameters are set the same as default.
Finetuning requires 2-8 nodes depending on the model size.

\myPara{Model Settings}
The model settings for \nameofmethod{}-D1 to \nameofmethod{}-D5 are listed in Table~\ref{tab:settings}.
We find that larger models (with 100M+ parameters) suffer overfitting.
To mitigate this issue, we set large stochastic depth rate for them.
Moreover, the learning rate selection also has a slight impact on the performance.
We find it is more beneficial to use larger initial learning rates for small-sized models.
In addition, the crop ratio can also slightly influence the performance.
Larger models prefer larger crop ratios.

\subsection{Main Results} \label{sec:main_results}

We compare the proposed \nameofmethod{} with the state-of-the-art models from the literature in Table~\ref{tab:sota}.
All results listed are based on using only ImageNet-1k images for training and no extra training data are used.
``Top-1,'' ``Real Top-1,'' and ``V2 Top-1'' refer to the top-1 accuracy using the original ImageNet validation labels,
cleaned-up real labels~\cite{beyer2020we}, and ImageNetV2 labels~\cite{recht2019imagenet}, respectively.
``Train size'' and ``Test size'' represent resolutions used in training and finetuning (test for CNNs).
We separate the results into five segments according to model size (number of parameters).

As can be seen, for different model sizes, our proposed \nameofmethod{} consistently performs better than previous state-of-the-art models.
Specially, taking the proposed \nameofmethod{}-D1 with 26.6M parameters as an example, 
testing on a resolution of 224 already yields 84.2\% top-1 accuracy on ImageNet.
Finetuning on 384 resolution further improves the performance to 85.2\%, which is clearly better than all the models with a comparable amount of training parameters.
When the model size is scaled up to 296M, we can achieve 87.1\% top-1 accuracy on ImageNet,
setting a new record in case of no extra training data.
\emph{To the best of our knowledge, our \nameofmethod{}-D5 is the first reaching 87.1\% top-1 accuracy
on ImageNet without extra training data}.

Our models also achieve the best results on the ``Real Top-1'' and ``V2 Top-1'' benchmarks.
As shown in Table~\ref{tab:sota}, our \nameofmethod{}-D4 with merely 193M parameters performs much better than previous state-of-the-art models, such as CaiT-M48 and NFNet.
Our models perform even better on the ImageNet-V2 benchmark.
As can be seen, our \nameofmethod{}-D3 can improve upon the previous
best result by 0.8\% (76.9\% \emph{v.s.} 77.7\%) using only a quarter of the parameters 
of CaiT-M48 (86M \emph{v.s.} 356M).
Our largest \nameofmethod{}-D5 can further boost the performance to 78\%.

\begin{table}[t]
  \centering
  \small
  \setlength\tabcolsep{1.3mm}
  \renewcommand\arraystretch{1}
  \caption{Ablation path from the LV-ViT-S~\cite{jiang2021token} baseline to our \nameofmethod{}-D1.
  All experiments, except for larger input resolution, can be finished within 3 days using a single server node with 8 V100 GPUs or 2 days with 8 A100 GPUs. Clearly, with only 27M learnable parameters, the performance
  can be boosted from 83.3 to 85.2 (\highlight{+1.9}) using the proposed \nameofmethod{} architecture. `T'
  and `O' refer to Transformer and \nameoflayer{}, respectively.}
  \label{tab:ablation_path}
  \begin{tabular}{lccccc} \toprule
    Training techniques & Layers & \#Param. & Top-1 Acc. (\%) \\ \midrule[0.5pt] 
    Baseline (LV-ViT-S \cite{jiang2021token})   & 16 & 26M & 83.3\\
    + Replace 2 Ts with Os                       & 16 & 25M & 83.7 (\highlight{+0.4}) \\
    + Add 2 more Os                             & 18 & 27M & 84.0 (\highlight{+0.7}) \\
    + \#Head in Ts ($6 \rightarrow 12$)         & 18 & 27M & 84.2 (\highlight{+0.9})\\
    + Resolution ($224 \rightarrow 384$)  & 18 & 27M & 85.2 (\highlight{+1.9})\\
    \bottomrule
  \end{tabular}
\end{table}

\subsection{Performance of \nameoflayer{}}
In this subsection, we demonstrate the importance of the proposed \nameoflayer{} in \nameofmethod{}.
We take the recent state-of-the-art vision transformer model, named LV-ViT-S, as our baseline.
LV-ViT-S contains 16 transformers in total and receives 83.3\% top-1 accuracy on ImageNet.
Each token in LV-ViT-S corresponds to an image patch of size $16\times16$, and hence
there are totally $14\times14$ tokens for a $224\times224$ input image.
The experiment path from the LV-ViT-S~\cite{jiang2021token} baseline to our \nameofmethod{}-D1
and the corresponding results can be found in Table~\ref{tab:ablation_path}.

As the goal of our proposed \nameoflayer{} is to encode expressive finer-level features, 
we first adjust the starting patch embedding module and change the patch size 
from $16\times16$ to $8\times8$.
We replace two transformers with our \nameoflayer{} at the fine level.
As can be seen from the second row of Table~\ref{tab:ablation_path},
such a slight adjustment brings us 0.4\% gain based on the baseline
that already reaches 83.3\% top-1 accuracy.
Adding another two \nameoflayer{}s further increases the performance to 83.9\%.
Finally, changing the head number in all the transformers from 6 to 12
and finetuning the resulting model at $384\times384$ resolution allows us to
yield a result of 85.2\%, which, to the best of our knowledge, is the first
time to attain 85+\% accuracy within less than 30M parameters.

\begin{table}[t]
  \centering
  \small
  \setlength\tabcolsep{2.5mm}
  \renewcommand\arraystretch{1}
  \caption{Performance of \nameoflayer{} against local self-attention and convolutions.
  For both self-attention and convolutions, we set the kernel size to $3\times3$.}
  
  \label{tab:outlooker_performance}
  \begin{tabular}{lccccc} \toprule
  Model & Layer type & \#Params  & Top-1 Acc.  \\ \midrule
  \nameofmethod{}-D1   & \nameoflayer{} & 27M & \highlight{84.2}   \\
  \nameofmethod{}-D1   & Local self-attention & 27M & 83.8  \\ 
  \nameofmethod{}-D1   & Convolution & 27M & 83.8   \\
    \bottomrule
  \end{tabular}
\end{table}

We also attempt to replace the proposed outlook attention with other methods for fine-level feature encoding, including local self-attention~\cite{liu2021swin} and spatial convolutions.
For a fair comparison, we set the window size to $3\times3$ for both local self-attention and convolutions.
The results can be found in Table~\ref{tab:outlooker_performance}.
As can be seen, under the same training recipe and architecture, our \nameoflayer{}
performs better than both local self-attention and convolutions.
In addition, we can also observe that local self-attention and convolutions
can also lift the performance compared to the LV-ViT-S baseline, demonstrating
that encoding fine-level token representations indeed helps.

\subsection{Ablation Analysis}

\myPara{Model Scaling}
We scale up the \nameofmethod{}-D1 model to 4 different models (\nameofmethod{}-D2 to \nameofmethod{}-D5)
in two different ways: 1) increasing the model size during training, including network depth, 
hidden dimension, expansion ratio in MLP, and head number in both Outlookers and Transformers, and 
2) increasing the image resolution during finetuning and test.
The specifications for all models have been shown in Table~\ref{tab:arch_details} and
their corresponding results can be found in Table~\ref{tab:model_scaling}.
We can observe that both aforementioned ways can largely improve the model performance.
From \nameofmethod{}-D1 to \nameofmethod{}-D2, there is 1\% improvement with doubled parameters.
Further increasing the model size form \nameofmethod{}-D2 to \nameofmethod{}-D5 yields nearly another
1\% accuracy gain.
In addition, for all the five models, increasing the resolution during finetuning brings around 1\% performance gain.

\begin{table}[t]
  \centering
  \small
  \setlength\tabcolsep{1.5mm}
  \renewcommand\arraystretch{1}
  \caption{Model performance when scaling up in two different ways: training model size and 
  testing resolution. The computations (M-Adds) reported here are based on $224\times224$
  image resolution.}
  
  \label{tab:model_scaling}
  \begin{tabular}{lccccc} \toprule
  Model & \#Params  & M-Adds & Top-1 Acc. &  Top-1 Acc.$\uparrow$ \\ \midrule[0.5pt]
  \nameofmethod{}-D1   & 26.6M & 6.8B & 84.2@224 & 85.4@384 &   \\
  \nameofmethod{}-D2   & 58.7M & 14.1B& 85.2@224 & 86.0@384 &   \\ 
  \nameofmethod{}-D3   & 86.3M & 20.6B& 85.4@224 & 86.3@448 &   \\
  \nameofmethod{}-D4   & 193M  & 43.8B& 85.7@224 &86.7@448 &   \\
  \nameofmethod{}-D5   & 296M  & 69.0B& \highlight{86.1@224}& \highlight{87.1@512} &   \\
    \bottomrule
  \end{tabular}
\end{table}

\myPara{Number of Outlookers} We observe that the number of \nameoflayer{}s used in our \nameofmethod{}
has an impact on the classification performance.
Here, we investigate the influence of using different numbers of \nameoflayer{}s
in our \nameofmethod{}.
Note that all \nameoflayer{}s act on finer-level token representations ($28\times28$).
The results have been shown in the top part of Table~\ref{tab:ablations}.
Without any \nameoflayer{}s, the baseline with 16 transformers receives 83.3\% accuracy.
Increasing the number of \nameoflayer{}s can improve the result
but the performance saturates when using 4 \nameoflayer{}s.
Further adding \nameoflayer{}s does not bring any performance gain.
Thus, when scaling up the model, we approximately use a ratio of 1:3 for
\nameoflayer{} and Transformers.

\myPara{Head Number in \nameoflayer{}s}
In Transformers, the channel dimension in each head is 
inversely proportional with the head number given a fixed hidden dimension.
Differently, in \nameoflayer{}s, the channel dimension in each head is fixed
when the kernel size is fixed (\ie, $C = K^4$).
So, will \nameoflayer{}s perform better if more heads are used?
In the bottom part of Table~\ref{tab:ablations}, we show the results with different head numbers
\nameoflayer{}s.
Experiments show that using more heads in \nameoflayer{}s can slightly improve
the performance with nearly no extra parameter increase but such increase stops when the head number is more than 6.
Therefore, by default, we set the head number in \nameoflayer{}s to 6 for 384 hidden dimension.
When the hidden dimension is set to 768, we use 12 heads in \nameoflayer{}s.

\begin{table}[t]
    \centering
    \small
    \setlength\tabcolsep{1.1mm}
    \renewcommand\arraystretch{1}
    \caption{More ablation experiments on \nameoflayer{}.
    `O' and `T' refer to \nameoflayer{} and Transformer, respectively.
    All results are based on \nameofmethod{}-D1 with test resolution $224\times224$.}
    \label{tab:ablations}
    \begin{tabular}{cccccccc} \toprule
    (\#O, \#T) & \#Heads in (O, T) & Kernel Size & \#Params & Top-1 Acc. \\ \toprule
    (0, 16)    & (-, 6) & $3\times3$  & 29.1M         & 83.3 \\
    (2, 14)    & (6, 6) & $3\times3$  & 25.9M         & 83.7 \\
    (4, 14)    & (6, 6) & $3\times3$  & 26.6M         & \highlight{84.0} \\
    (6, 12)    & (6, 6) & $3\times3$  & 24.5M         & 83.9 \\ \toprule
    (4, 14)    & (2, 6)            & $3\times3$  & 26.4M         & 83.9 \\ 
    (4, 14)    & (4, 6)            & $3\times3$  & 26.5M         & 83.9 \\
    (4, 14)    & (6, 6)            & $3\times3$  & 26.6M         & 84.0 \\
    (4, 14)    & (8, 6)            & $3\times3$  & 26.8M         & 84.0 \\
    (4, 14)    & (6, 12)           & $3\times3$  & 26.6M         & \highlight{84.2} \\ 
    \bottomrule
    \end{tabular}
\end{table}

\subsection{Semantic Segmentation}

In this subsection, we use our \nameofmethod{} as pretrained models
to evaluate the performance in semantic segmentation.
Our code is based on \texttt{mmsegmentation}~\cite{mmseg2020}.
We report results on two widely-used segmentation benchmarks:
Cityscapes~\cite{cordts2016cityscapes} and ADE20K~\cite{zhou2019semantic}.
The UperNet \cite{xiao2018unified} segmentation framework is adopted.
In training, we utilize the AdamW optimizer with an initial learning rate 
of 6e-5 and a weight decay of 0.01.
We also use a linear learning schedule with a minimum learning rate of 5e-6.
All models can be trained on a machine node with 8 A100 GPUs.
For cityscapes, we set the batch size to 8 and the input resolution to $1024\times1024$.
For ADE20K, the batch size is set to 16 and input resolution $512\times512$ is used.
As suggested by \cite{zhou2019semantic}, we report results in terms of mean intersection-over-union (mIoU) for both datasets and mean pixel accuracy for
ADE20K.
In inference, we perform multi-scale test with interpolation rates of 
[0.75, 1.0, 1.25, 1.5, 1.75].

\subsubsection{Cityscapes}

Cityscapes \cite{cordts2016cityscapes} is one of the most popular datasets 
for semantic segmentation, which targets at street scene segmentation.
It has 5K high-quality pixel-annotated images with resolution $1024\times2048$
and contains 19 classes in total.
As in most previous work, we split the whole dataset into three splits 
for training, validation and test, which contain 2,975, 500, and 1,525 images, respectively.
We report results on the validation set.
The comparison results can be found in Table~\ref{tab:val_city}.
It is obvious that the proposed approach outperforms all other methods,
including the recent state-of-the-art SegFormer-B5 model.
Our \nameofmethod{}-D4 with UperNet decoder head achieves the best result
84.3\%, 0.3\% better than the previous state-of-the-art result 84.0\% made
by SegFormer-B5.
According to PaperWithCode\footnote{\url{https://paperswithcode.com/sota/semantic-segmentation-on-cityscapes-val}}, this is a new state-of-the-art result on Cityscapes validation set.

\begin{table}[t]
  \centering
  \small
  \setlength\tabcolsep{1.5mm}
  \renewcommand\arraystretch{1.0}
  \caption{Comparisons with the state-of-the-arts on the Cityscapes validation set \cite{cordts2016cityscapes}. `Pretrained' refers to the dataset each backbone network is pretrained on. All models are trained on the training set and multi-scale test results are reported in the `mIoU' column.}
  \begin{tabular}{lccc} \toprule
    Method &  Backbone & Pretrained & mIoU \\ \toprule
    DenseASPP \cite{yang2018denseaspp}  & DenseNet~\cite{huang2017densely} & ImgNet-1k & 80.6  \\
    DeepLabv3+ \cite{chen2018encoder} & Xception-65~\cite{chollet2017xception} & ImgNet-1k & 79.1 \\
    DPC \cite{chen2018searching} & Xception-71~\cite{chollet2017xception} & ImgNet-1k & 80.8 \\
    DANet \cite{fu2019dual}  & ResNet-101 & ImgNet-1k &  81.5  \\
    CCNet \cite{huang2018ccnet} & ResNet-101 & ImgNet-1k & 81.3 \\ 
    Strip Pooling~\cite{hou2020strip} & ResNet-101 & ImgNet-1k & 81.9 \\ \midrule
    
    SETR \cite{zheng2020rethinking} & ViT-L~\cite{dosovitskiy2020image} & ImgNet-22k &  82.1 \\
    PatchDiverse~\cite{gong2021improve} & Swin-L~\cite{liu2021swin} & ImgNet-22k & 83.6 \\
    SpineNet-S143+~\cite{rashwan2021dilated} & SpineNet & ImgNet-1k & 83.0 \\
    SegFormer-B5~\cite{xie2021segformer} & SegFormer & ImgNet-1k & 84.0 \\
    \midrule
    \nameofmethod{}-D1 & \nameofmethod{} & ImgNet-1k &  83.1 \\
    \nameofmethod{}-D4 & \nameofmethod{} & ImgNet-1k &  \highlight{84.3} \\
    \bottomrule
  \end{tabular}
  \label{tab:val_city}
\end{table}

\subsubsection{ADE20K}
We also run experiments on the widely-used ADE20K \cite{zhou2019semantic} dataset. 
ADE20K contains 25K images in total, including 20K images for training, 2K images for validation, and 3K images for test.
It covers 150 different common foreground categories.
We compare our segmentation results with previous state-of-the-art 
segmentation methods in Table~\ref{tab:seg_comp}.
Without pretraining on large-scale datasets, such as ImageNet-22K, 
our \nameofmethod{}-D1 with UperNet achieves an mIoU score of 50.5.
When the \nameofmethod{}-D5 is used as backbone,
the mIoU score can be further improved to 54.3, a new state-of-the-art result
on ADE20K with no extra pretraining data except for ImageNet-1k.

\begin{table}[h]
  \centering
  \small
  \setlength\tabcolsep{1.2mm}
  \renewcommand\arraystretch{1.0}
  \caption{Comparison with previous state-of-the-art methods on the ADE20K validation set. Our \nameofmethod{}-D5 achieves the best result on ADE20K with only ImageNet-1K
  as training data in pretraining. `Pixel' refers to mean pixel accuracy.}
  \label{tab:seg_comp}
  \begin{tabular}{lcccccccc} \toprule
    Method & Backbone & Pretrained & mIoU & Pixel \\ \toprule
    PSPNet \cite{zhao2017pyramid} & ResNet-269 & ImgNet-1k  & 44.9 & 81.7 \\
    UperNet \cite{xiao2018unified} & ResNet-101 & ImgNet-1k & 44.9 & - \\
    Strip Pooling \cite{hou2020strip} & ResNet-101 & ImgNet-1k & 45.6 & 82.1\\
    DeepLabV3+ \cite{chen2018encoder} & ResNeSt200 & ImgNet-1k & 48.4& -  \\ 
    \midrule[0.5pt]
    SETR \cite{zheng2020rethinking} & ViT-Large & ImgNet-22k & 50.3 &  83.5 \\
    SegFormer-B5~\cite{xie2021segformer} & SegFormer & ImgNet-1k & 51.8 & - \\
    Swin-B \cite{liu2021swin} & Swin-B & ImgNet-22k & 51.6 & - \\
    Seg-L-Mask/16~\cite{strudel2021segmenter} & ViT-Large & ImgNet-22k & 53.2 & - \\
    Swin-L \cite{liu2021swin} & Swin-L & ImgNet-22k & 53.5 & - \\\midrule[0.5pt]
    \nameofmethod{}-D1 & \nameofmethod{} & ImgNet-1k   & 50.5 & 83.3 \\
    \nameofmethod{}-D3 & \nameofmethod{} &ImgNet-1k   & 52.9 & 84.6 \\
    \nameofmethod{}-D5 & \nameofmethod{} &ImgNet-1k   & \highlight{54.3} & \highlight{85.0} \\
    \bottomrule
  \end{tabular}
\end{table}

\section{Related Work}

As one of the most fundamental problems in computer vision, image classification
has experienced remarkable progress since the introduction of deep neural network models.
In what follows, we briefly review those successful models that are closely related to this work.

Earlier models attaining state-of-the-art performance for image classification 
are mostly CNN-based ones that simply stack a sequence of spatial convolutions and poolings, 
represented by AlexNet~\cite{krizhevsky2012imagenet} and VGGNet \cite{simonyan2014very}.
ResNets~\cite{he2016deep} advances the design of CNN architectures by introducing skip connections
to enable training of very deep models.
Inceptions~\cite{szegedy2015going,szegedy2016rethinking,szegedy2017inception}
and ResNeXt \cite{xie2017aggregated} examine the design principles of the model building blocks and 
introduce multiple parallel paths of sets of specialized filters.
SENet~\cite{hu2018squeeze} presents a squeeze-and-excitation module to explicitly
model the inter-dependencies among channels.
DPNs~\cite{chen2017dual} leverage both residual and dense connections
for designing stronger building blocks.
EfficientNet~\cite{tan2019efficientnet} and NasNet~\cite{zoph2018learning} take
advantage of neural architecture search to search for powerful network architectures.
Later state-of-the-art models~\cite{huang2019gpipe,touvron2019fixing,xie2020adversarial} 
mostly utilize different training or optimization methods or finetuning techniques to
improve EfficientNet.
Very recently, NFNet~\cite{brock2021high} breaks the dominance of EfficientNet
by designing a normalization-free architecture, making the first work attaining 86.5\% top-1 accuracy
on ImageNet using no extra data.
CNNs, as the de-facto networks in visual recognition for years,
have indeed been very  successful but their focus is on how to learn more discriminative 
local features by designing better architectures.
Essentially, they are short of the capability of explicitly building global relationships
among representations that have been proven crucial \cite{wang2018non}.

Recent progress on image classification is mostly driven by attention-based models \cite{zhao2020exploring,wang2018non,hu2019local} or specifically transformer-based models.
Transformers make use of the self-attention mechanism, making modeling long-range dependencies
possible.
Transformers~\cite{vaswani2017attention} are originally designed for natural language tasks~\cite{devlin2018bert,radford2018improving,brown2020language,yang2019xlnet,peters2019knowledge,liu2019roberta} and have recently been demonstrated  effective in image classification.
Dosovitskiy \emph{et al.}~\cite{dosovitskiy2020image} are among the first to show that purely transformer-based architectures (\ie, ViT) can also
get state-of-the-art performance in image classification but require large-scale datasets,
such as ImageNet-22k and JFT-300M (which is not publicly available) for pretraining. 
DeiT~\cite{touvron2020training} and T2T-ViT \cite{yuan2021tokens} mitigate the problem of ViTs
requiring large-scale datasets and propose data efficient ViTs.
Since then, a surge of works on ViT continuously come into being with further improvements.
Some of them~\cite{chen2021crossvit,han2021transformer,wu2021cvt,vaswani2021scaling,zhou2021refiner} 
introduce local dependency into vision transformers by modifying the patch embedding block 
or the transformer block or both, 
while others~\cite{heo2021rethinking,liu2021swin,wang2021pyramid} adopt 
a pyramid structure to reduce the overall computation while maintaining the models' ability 
to capture low-level features.
There are also some works \cite{zhou2021deepvit,zhai2021scaling,touvron2021going,gong2021improve} 
aiming at solving the optimization and scaling problems of ViTs.

Our \nameofmethod{}  not only models long-range dependencies 
but also encodes fine-level features into token representations
by the proposed  \nameoflayer{}.
Unlike the recent  hybrid architectures (\emph{e.g.}, Hybrid-ViT \cite{dosovitskiy2020image} and
BoTNet~\cite{srinivas2021bottleneck}) that rely on convolutions for  feature encoding,
\nameoflayer{} proposes to use local pair-wise token similarities   to encode fine-level features and spatial context into tokens features  and 
hence is more effective and parameter-efficient. 
This also makes our model different from the Dynamic Convolution~\cite{wu2019pay} and Involution~\cite{li2021involution} that  generate
input-dependent convolution kernels to encode the features.

\section{Conclusions}
We presented a new model, Vision Outlooker (VOLO). Extensive experiments for image classification and segmentation demonstrate VOLO outperforms CNN- and Transformer-based models, and establishes new SOTA results. We hope that the strong performance of VOLO on several computer vision tasks will encourage follow-up research on better fine-level feature learning.  The performance superiority of VOLO comes from the new outlook attention mechanism that dynamically aggregates fine-level features in a dense manner, and we will continue our investigation in other applications, like natural language processing.

\section*{Acknowledgement}

We gratefully acknowledge the support of NVIDIA AI Tech Center (NVAITC) to this research project, especially the great helps in GPU technology supports from Terry Jianxiong Yin (NVAITC) and Qingyi Tao (NVAITC). 

\medskip

{
\small
\bibliographystyle{plain}
\bibliography{ref}

\begin{thebibliography}{10}

\bibitem{beyer2020we}
Lucas Beyer, Olivier~J H{\'e}naff, Alexander Kolesnikov, Xiaohua Zhai, and
  A{\"a}ron van~den Oord.
\newblock Are we done with imagenet?
\newblock {\em arXiv preprint arXiv:2006.07159}, 2020.

\bibitem{brock2021high}
Andrew Brock, Soham De, Samuel~L Smith, and Karen Simonyan.
\newblock High-performance large-scale image recognition without normalization.
\newblock {\em arXiv preprint arXiv:2102.06171}, 2021.

\bibitem{brown2020language}
Tom~B Brown, Benjamin Mann, Nick Ryder, Melanie Subbiah, Jared Kaplan, Prafulla
  Dhariwal, Arvind Neelakantan, Pranav Shyam, Girish Sastry, Amanda Askell,
  et~al.
\newblock Language models are few-shot learners.
\newblock {\em arXiv preprint arXiv:2005.14165}, 2020.

\bibitem{chen2021crossvit}
Chun-Fu Chen, Quanfu Fan, and Rameswar Panda.
\newblock Crossvit: Cross-attention multi-scale vision transformer for image
  classification.
\newblock {\em arXiv preprint arXiv:2103.14899}, 2021.

\bibitem{chen2018searching}
Liang-Chieh Chen, Maxwell Collins, Yukun Zhu, George Papandreou, Barret Zoph,
  Florian Schroff, Hartwig Adam, and Jon Shlens.
\newblock Searching for efficient multi-scale architectures for dense image
  prediction.
\newblock In {\em NeurIPS}, pages 8699--8710, 2018.

\bibitem{chen2018encoder}
Liang-Chieh Chen, Yukun Zhu, George Papandreou, Florian Schroff, and Hartwig
  Adam.
\newblock Encoder-decoder with atrous separable convolution for semantic image
  segmentation.
\newblock In {\em Proceedings of the European conference on computer vision
  (ECCV)}, pages 801--818, 2018.

\bibitem{chen2017dual}
Yunpeng Chen, Jianan Li, Huaxin Xiao, Xiaojie Jin, Shuicheng Yan, and Jiashi
  Feng.
\newblock Dual path networks.
\newblock In {\em Advances in Neural Information Processing Systems}, pages
  4467--4475, 2017.

\bibitem{chollet2017xception}
Fran{\c{c}}ois Chollet.
\newblock Xception: Deep learning with depthwise separable convolutions.
\newblock In {\em Proceedings of the IEEE conference on computer vision and
  pattern recognition}, pages 1251--1258, 2017.

\bibitem{mmseg2020}
MMSegmentation Contributors.
\newblock {MMSegmentation}: Openmmlab semantic segmentation toolbox and
  benchmark.
\newblock \url{https://github.com/open-mmlab/mmsegmentation}, 2020.

\bibitem{cordts2016cityscapes}
Marius Cordts, Mohamed Omran, Sebastian Ramos, Timo Rehfeld, Markus Enzweiler,
  Rodrigo Benenson, Uwe Franke, Stefan Roth, and Bernt Schiele.
\newblock The cityscapes dataset for semantic urban scene understanding.
\newblock In {\em Proceedings of the IEEE conference on computer vision and
  pattern recognition}, 2016.

\bibitem{cubuk2020randaugment}
Ekin~D Cubuk, Barret Zoph, Jonathon Shlens, and Quoc~V Le.
\newblock Randaugment: Practical automated data augmentation with a reduced
  search space.
\newblock In {\em Proceedings of the IEEE/CVF Conference on Computer Vision and
  Pattern Recognition Workshops}, pages 702--703, 2020.

\bibitem{deng2009imagenet}
Jia Deng, Wei Dong, Richard Socher, Li-Jia Li, Kai Li, and Li~Fei-Fei.
\newblock Imagenet: A large-scale hierarchical image database.
\newblock In {\em 2009 IEEE conference on computer vision and pattern
  recognition}, pages 248--255. Ieee, 2009.

\bibitem{devlin2018bert}
Jacob Devlin, Ming-Wei Chang, Kenton Lee, and Kristina Toutanova.
\newblock Bert: Pre-training of deep bidirectional transformers for language
  understanding.
\newblock {\em arXiv preprint arXiv:1810.04805}, 2018.

\bibitem{dosovitskiy2020image}
Alexey Dosovitskiy, Lucas Beyer, Alexander Kolesnikov, Dirk Weissenborn,
  Xiaohua Zhai, Thomas Unterthiner, Mostafa Dehghani, Matthias Minderer, Georg
  Heigold, Sylvain Gelly, et~al.
\newblock An image is worth 16x16 words: Transformers for image recognition at
  scale.
\newblock {\em arXiv preprint arXiv:2010.11929}, 2020.

\bibitem{foret2020sharpness}
Pierre Foret, Ariel Kleiner, Hossein Mobahi, and Behnam Neyshabur.
\newblock Sharpness-aware minimization for efficiently improving
  generalization.
\newblock {\em arXiv preprint arXiv:2010.01412}, 2020.

\bibitem{fort2021drawing}
Stanislav Fort, Andrew Brock, Razvan Pascanu, Soham De, and Samuel~L Smith.
\newblock Drawing multiple augmentation samples per image during training
  efficiently decreases test error.
\newblock {\em arXiv preprint arXiv:2105.13343}, 2021.

\bibitem{fu2019dual}
Jun Fu, Jing Liu, Haijie Tian, Yong Li, Yongjun Bao, Zhiwei Fang, and Hanqing
  Lu.
\newblock Dual attention network for scene segmentation.
\newblock In {\em Proceedings of the IEEE Conference on Computer Vision and
  Pattern Recognition}, pages 3146--3154, 2019.

\bibitem{gong2021improve}
Chengyue Gong, Dilin Wang, Meng Li, Vikas Chandra, and Qiang Liu.
\newblock Improve vision transformers training by suppressing over-smoothing.
\newblock {\em arXiv preprint arXiv:2104.12753}, 2021.

\bibitem{han2021transformer}
Kai Han, An~Xiao, Enhua Wu, Jianyuan Guo, Chunjing Xu, and Yunhe Wang.
\newblock Transformer in transformer.
\newblock {\em arXiv preprint arXiv:2103.00112}, 2021.

\bibitem{he2016deep}
Kaiming He, Xiangyu Zhang, Shaoqing Ren, and Jian Sun.
\newblock Deep residual learning for image recognition.
\newblock In {\em Proceedings of the IEEE conference on computer vision and
  pattern recognition}, pages 770--778, 2016.

\bibitem{heo2021rethinking}
Byeongho Heo, Sangdoo Yun, Dongyoon Han, Sanghyuk Chun, Junsuk Choe, and
  Seong~Joon Oh.
\newblock Rethinking spatial dimensions of vision transformers.
\newblock {\em arXiv preprint arXiv:2103.16302}, 2021.

\bibitem{hinton2015distilling}
Geoffrey Hinton, Oriol Vinyals, and Jeff Dean.
\newblock Distilling the knowledge in a neural network.
\newblock {\em arXiv preprint arXiv:1503.02531}, 2015.

\bibitem{hou2021vision}
Qibin Hou, Zihang Jiang, Li~Yuan, Ming-Ming Cheng, Shuicheng Yan, and Jiashi
  Feng.
\newblock Vision permutator: A permutable mlp-like architecture for visual
  recognition, 2021.

\bibitem{hou2020strip}
Qibin Hou, Li~Zhang, Ming-Ming Cheng, and Jiashi Feng.
\newblock Strip pooling: Rethinking spatial pooling for scene parsing.
\newblock In {\em Proceedings of the IEEE/CVF Conference on Computer Vision and
  Pattern Recognition}, pages 4003--4012, 2020.

\bibitem{hou2021coordinate}
Qibin Hou, Daquan Zhou, and Jiashi Feng.
\newblock Coordinate attention for efficient mobile network design.
\newblock In {\em Proceedings of the IEEE conference on computer vision and
  pattern recognition}, 2021.

\bibitem{hu2019local}
Han Hu, Zheng Zhang, Zhenda Xie, and Stephen Lin.
\newblock Local relation networks for image recognition.
\newblock In {\em Proceedings of the IEEE International Conference on Computer
  Vision}, pages 3464--3473, 2019.

\bibitem{hu2018squeeze}
Jie Hu, Li~Shen, and Gang Sun.
\newblock Squeeze-and-excitation networks.
\newblock In {\em Proceedings of the IEEE conference on computer vision and
  pattern recognition}, pages 7132--7141, 2018.

\bibitem{huang2017densely}
Gao Huang, Zhuang Liu, Laurens Van Der~Maaten, and Kilian~Q Weinberger.
\newblock Densely connected convolutional networks.
\newblock In {\em Proceedings of the IEEE conference on computer vision and
  pattern recognition}, pages 4700--4708, 2017.

\bibitem{huang2016deep}
Gao Huang, Yu~Sun, Zhuang Liu, Daniel Sedra, and Kilian~Q Weinberger.
\newblock Deep networks with stochastic depth.
\newblock In {\em European conference on computer vision}, pages 646--661.
  Springer, 2016.

\bibitem{huang2019gpipe}
Yanping Huang, Youlong Cheng, Ankur Bapna, Orhan Firat, Dehao Chen, Mia Chen,
  HyoukJoong Lee, Jiquan Ngiam, Quoc~V Le, Yonghui Wu, et~al.
\newblock Gpipe: Efficient training of giant neural networks using pipeline
  parallelism.
\newblock {\em Advances in neural information processing systems}, 32:103--112,
  2019.

\bibitem{huang2018ccnet}
Zilong Huang, Xinggang Wang, Lichao Huang, Chang Huang, Yunchao Wei, and Wenyu
  Liu.
\newblock Ccnet: Criss-cross attention for semantic segmentation.
\newblock {\em arXiv preprint arXiv:1811.11721}, 2018.

\bibitem{jiang2021token}
Zihang Jiang, Qibin Hou, Li~Yuan, Daquan Zhou, Xiaojie Jin, Anran Wang, and
  Jiashi Feng.
\newblock All tokens matter: Token labeling for training better vision
  transformers.
\newblock {\em arXiv preprint arXiv:2104.10858}, 2021.

\bibitem{krizhevsky2012imagenet}
Alex Krizhevsky, Ilya Sutskever, and Geoffrey~E Hinton.
\newblock Imagenet classification with deep convolutional neural networks.
\newblock {\em Advances in neural information processing systems},
  25:1097--1105, 2012.

\bibitem{li2021involution}
Duo Li, Jie Hu, Changhu Wang, Xiangtai Li, Qi~She, Lei Zhu, Tong Zhang, and
  Qifeng Chen.
\newblock Involution: Inverting the inherence of convolution for visual
  recognition.
\newblock In {\em Proceedings of the IEEE/CVF Conference on Computer Vision and
  Pattern Recognition}, pages 12321--12330, 2021.

\bibitem{liu2020rethinking}
Fenglin Liu, Xuancheng Ren, Zhiyuan Zhang, Xu~Sun, and Yuexian Zou.
\newblock Rethinking skip connection with layer normalization.
\newblock In {\em Proceedings of the 28th International Conference on
  Computational Linguistics}, pages 3586--3598, 2020.

\bibitem{liu2019roberta}
Yinhan Liu, Myle Ott, Naman Goyal, Jingfei Du, Mandar Joshi, Danqi Chen, Omer
  Levy, Mike Lewis, Luke Zettlemoyer, and Veselin Stoyanov.
\newblock Roberta: A robustly optimized bert pretraining approach.
\newblock {\em arXiv preprint arXiv:1907.11692}, 2019.

\bibitem{liu2021swin}
Ze~Liu, Yutong Lin, Yue Cao, Han Hu, Yixuan Wei, Zheng Zhang, Stephen Lin, and
  Baining Guo.
\newblock Swin transformer: Hierarchical vision transformer using shifted
  windows.
\newblock {\em arXiv preprint arXiv:2103.14030}, 2021.

\bibitem{loshchilov2017decoupled}
Ilya Loshchilov and Frank Hutter.
\newblock Decoupled weight decay regularization.
\newblock {\em arXiv preprint arXiv:1711.05101}, 2017.

\bibitem{paszke2019pytorch}
Adam Paszke, Sam Gross, Francisco Massa, Adam Lerer, James Bradbury, Gregory
  Chanan, Trevor Killeen, Zeming Lin, Natalia Gimelshein, Luca Antiga, et~al.
\newblock Pytorch: An imperative style, high-performance deep learning library.
\newblock In {\em Advances in neural information processing systems}, pages
  8026--8037, 2019.

\bibitem{peters2019knowledge}
Matthew~E Peters, Mark Neumann, Robert~L Logan~IV, Roy Schwartz, Vidur Joshi,
  Sameer Singh, and Noah~A Smith.
\newblock Knowledge enhanced contextual word representations.
\newblock {\em arXiv preprint arXiv:1909.04164}, 2019.

\bibitem{radford2018improving}
Alec Radford, Karthik Narasimhan, Tim Salimans, and Ilya Sutskever.
\newblock Improving language understanding by generative pre-training, 2018.

\bibitem{rashwan2021dilated}
Abdullah Rashwan, Xianzhi Du, Xiaoqi Yin, and Jing Li.
\newblock Dilated spinenet for semantic segmentation.
\newblock {\em arXiv preprint arXiv:2103.12270}, 2021.

\bibitem{recht2019imagenet}
Benjamin Recht, Rebecca Roelofs, Ludwig Schmidt, and Vaishaal Shankar.
\newblock Do imagenet classifiers generalize to imagenet?
\newblock In {\em International Conference on Machine Learning}, pages
  5389--5400. PMLR, 2019.

\bibitem{simonyan2014very}
Karen Simonyan and Andrew Zisserman.
\newblock Very deep convolutional networks for large-scale image recognition.
\newblock {\em arXiv preprint arXiv:1409.1556}, 2014.

\bibitem{srinivas2021bottleneck}
Aravind Srinivas, Tsung-Yi Lin, Niki Parmar, Jonathon Shlens, Pieter Abbeel,
  and Ashish Vaswani.
\newblock Bottleneck transformers for visual recognition.
\newblock {\em arXiv preprint arXiv:2101.11605}, 2021.

\bibitem{strudel2021segmenter}
Robin Strudel, Ricardo Garcia, Ivan Laptev, and Cordelia Schmid.
\newblock Segmenter: Transformer for semantic segmentation.
\newblock {\em arXiv preprint arXiv:2105.05633}, 2021.

\bibitem{szegedy2017inception}
Christian Szegedy, Sergey Ioffe, Vincent Vanhoucke, and Alexander~A Alemi.
\newblock Inception-v4, inception-resnet and the impact of residual connections
  on learning.
\newblock In {\em AAAI}, volume~4, page~12, 2017.

\bibitem{szegedy2015going}
Christian Szegedy, Wei Liu, Yangqing Jia, Pierre Sermanet, Scott Reed, Dragomir
  Anguelov, Dumitru Erhan, Vincent Vanhoucke, and Andrew Rabinovich.
\newblock Going deeper with convolutions.
\newblock In {\em Proceedings of the IEEE conference on computer vision and
  pattern recognition}, pages 1--9, 2015.

\bibitem{szegedy2016rethinking}
Christian Szegedy, Vincent Vanhoucke, Sergey Ioffe, Jon Shlens, and Zbigniew
  Wojna.
\newblock Rethinking the inception architecture for computer vision.
\newblock In {\em Proceedings of the IEEE conference on computer vision and
  pattern recognition}, pages 2818--2826, 2016.

\bibitem{tan2019efficientnet}
Mingxing Tan and Quoc~V Le.
\newblock Efficientnet: Rethinking model scaling for convolutional neural
  networks.
\newblock {\em arXiv preprint arXiv:1905.11946}, 2019.

\bibitem{touvron2020training}
Hugo Touvron, Matthieu Cord, Matthijs Douze, Francisco Massa, Alexandre
  Sablayrolles, and Herv{\'e} J{\'e}gou.
\newblock Training data-efficient image transformers \& distillation through
  attention.
\newblock {\em arXiv preprint arXiv:2012.12877}, 2020.

\bibitem{touvron2021going}
Hugo Touvron, Matthieu Cord, Alexandre Sablayrolles, Gabriel Synnaeve, and
  Herv{\'e} J{\'e}gou.
\newblock Going deeper with image transformers.
\newblock {\em arXiv preprint arXiv:2103.17239}, 2021.

\bibitem{touvron2019fixing}
Hugo Touvron, Andrea Vedaldi, Matthijs Douze, and Herv{\'e} J{\'e}gou.
\newblock Fixing the train-test resolution discrepancy.
\newblock {\em arXiv preprint arXiv:1906.06423}, 2019.

\bibitem{vaswani2021scaling}
Ashish Vaswani, Prajit Ramachandran, Aravind Srinivas, Niki Parmar, Blake
  Hechtman, and Jonathon Shlens.
\newblock Scaling local self-attention for parameter efficient visual
  backbones.
\newblock {\em arXiv preprint arXiv:2103.12731}, 2021.

\bibitem{vaswani2017attention}
Ashish Vaswani, Noam Shazeer, Niki Parmar, Jakob Uszkoreit, Llion Jones,
  Aidan~N Gomez, {\L}ukasz Kaiser, and Illia Polosukhin.
\newblock Attention is all you need.
\newblock {\em Advances in neural information processing systems},
  30:5998--6008, 2017.

\bibitem{wang2020axial}
Huiyu Wang, Yukun Zhu, Bradley Green, Hartwig Adam, Alan Yuille, and
  Liang-Chieh Chen.
\newblock Axial-deeplab: Stand-alone axial-attention for panoptic segmentation.
\newblock In {\em European Conference on Computer Vision}, pages 108--126.
  Springer, 2020.

\bibitem{wang2021pyramid}
Wenhai Wang, Enze Xie, Xiang Li, Deng-Ping Fan, Kaitao Song, Ding Liang, Tong
  Lu, Ping Luo, and Ling Shao.
\newblock Pyramid vision transformer: A versatile backbone for dense prediction
  without convolutions.
\newblock {\em arXiv preprint arXiv:2102.12122}, 2021.

\bibitem{wang2018non}
Xiaolong Wang, Ross Girshick, Abhinav Gupta, and Kaiming He.
\newblock Non-local neural networks.
\newblock In {\em Proceedings of the IEEE conference on computer vision and
  pattern recognition}, pages 7794--7803, 2018.

\bibitem{rw2019timm}
Ross Wightman.
\newblock Pytorch image models.
\newblock \url{https://github.com/rwightman/pytorch-image-models}, 2019.

\bibitem{wu2019pay}
Felix Wu, Angela Fan, Alexei Baevski, Yann~N Dauphin, and Michael Auli.
\newblock Pay less attention with lightweight and dynamic convolutions.
\newblock {\em arXiv preprint arXiv:1901.10430}, 2019.

\bibitem{wu2021cvt}
Haiping Wu, Bin Xiao, Noel Codella, Mengchen Liu, Xiyang Dai, Lu~Yuan, and Lei
  Zhang.
\newblock Cvt: Introducing convolutions to vision transformers.
\newblock {\em arXiv preprint arXiv:2103.15808}, 2021.

\bibitem{xiao2018unified}
Tete Xiao, Yingcheng Liu, Bolei Zhou, Yuning Jiang, and Jian Sun.
\newblock Unified perceptual parsing for scene understanding.
\newblock In {\em Proceedings of the European Conference on Computer Vision
  (ECCV)}, pages 418--434, 2018.

\bibitem{xie2020adversarial}
Cihang Xie, Mingxing Tan, Boqing Gong, Jiang Wang, Alan~L Yuille, and Quoc~V
  Le.
\newblock Adversarial examples improve image recognition.
\newblock In {\em Proceedings of the IEEE/CVF Conference on Computer Vision and
  Pattern Recognition}, pages 819--828, 2020.

\bibitem{xie2021segformer}
Enze Xie, Wenhai Wang, Zhiding Yu, Anima Anandkumar, Jose~M Alvarez, and Ping
  Luo.
\newblock Segformer: Simple and efficient design for semantic segmentation with
  transformers.
\newblock {\em arXiv preprint arXiv:2105.15203}, 2021.

\bibitem{xie2017aggregated}
Saining Xie, Ross Girshick, Piotr Doll{\'a}r, Zhuowen Tu, and Kaiming He.
\newblock Aggregated residual transformations for deep neural networks.
\newblock In {\em Proceedings of the IEEE conference on computer vision and
  pattern recognition}, pages 1492--1500, 2017.

\bibitem{yang2018denseaspp}
Maoke Yang, Kun Yu, Chi Zhang, Zhiwei Li, and Kuiyuan Yang.
\newblock Denseaspp for semantic segmentation in street scenes.
\newblock In {\em Proceedings of the IEEE conference on computer vision and
  pattern recognition}, 2018.

\bibitem{yang2019xlnet}
Zhilin Yang, Zihang Dai, Yiming Yang, Jaime Carbonell, Russ~R Salakhutdinov,
  and Quoc~V Le.
\newblock Xlnet: Generalized autoregressive pretraining for language
  understanding.
\newblock In {\em Advances in neural information processing systems}, pages
  5753--5763, 2019.

\bibitem{yuan2021tokens}
Li~Yuan, Yunpeng Chen, Tao Wang, Weihao Yu, Yujun Shi, Francis~EH Tay, Jiashi
  Feng, and Shuicheng Yan.
\newblock Tokens-to-token vit: Training vision transformers from scratch on
  imagenet.
\newblock {\em arXiv preprint arXiv:2101.11986}, 2021.

\bibitem{yuan2020revisiting}
Li~Yuan, Francis~EH Tay, Guilin Li, Tao Wang, and Jiashi Feng.
\newblock Revisiting knowledge distillation via label smoothing regularization.
\newblock In {\em Proceedings of the IEEE/CVF Conference on Computer Vision and
  Pattern Recognition}, pages 3903--3911, 2020.

\bibitem{yun2019cutmix}
Sangdoo Yun, Dongyoon Han, Seong~Joon Oh, Sanghyuk Chun, Junsuk Choe, and
  Youngjoon Yoo.
\newblock Cutmix: Regularization strategy to train strong classifiers with
  localizable features.
\newblock In {\em Proceedings of the IEEE/CVF International Conference on
  Computer Vision}, pages 6023--6032, 2019.

\bibitem{zhai2021scaling}
Xiaohua Zhai, Alexander Kolesnikov, Neil Houlsby, and Lucas Beyer.
\newblock Scaling vision transformers.
\newblock {\em arXiv preprint arXiv:2106.04560}, 2021.

\bibitem{zhang2017mixup}
Hongyi Zhang, Moustapha Cisse, Yann~N Dauphin, and David Lopez-Paz.
\newblock mixup: Beyond empirical risk minimization.
\newblock {\em arXiv preprint arXiv:1710.09412}, 2017.

\bibitem{zhao2020exploring}
Hengshuang Zhao, Jiaya Jia, and Vladlen Koltun.
\newblock Exploring self-attention for image recognition.
\newblock In {\em Proceedings of the IEEE/CVF Conference on Computer Vision and
  Pattern Recognition}, pages 10076--10085, 2020.

\bibitem{zhao2017pyramid}
Hengshuang Zhao, Jianping Shi, Xiaojuan Qi, Xiaogang Wang, and Jiaya Jia.
\newblock Pyramid scene parsing network.
\newblock In {\em Proceedings of the IEEE conference on computer vision and
  pattern recognition}, pages 2881--2890, 2017.

\bibitem{zheng2020rethinking}
Sixiao Zheng, Jiachen Lu, Hengshuang Zhao, Xiatian Zhu, Zekun Luo, Yabiao Wang,
  Yanwei Fu, Jianfeng Feng, Tao Xiang, Philip~HS Torr, et~al.
\newblock Rethinking semantic segmentation from a sequence-to-sequence
  perspective with transformers.
\newblock {\em arXiv preprint arXiv:2012.15840}, 2020.

\bibitem{zhong2020random}
Zhun Zhong, Liang Zheng, Guoliang Kang, Shaozi Li, and Yi~Yang.
\newblock Random erasing data augmentation.
\newblock In {\em Proceedings of the AAAI Conference on Artificial
  Intelligence}, volume~34, pages 13001--13008, 2020.

\bibitem{zhou2019semantic}
Bolei Zhou, Hang Zhao, Xavier Puig, Tete Xiao, Sanja Fidler, Adela Barriuso,
  and Antonio Torralba.
\newblock Semantic understanding of scenes through the ade20k dataset.
\newblock {\em International Journal of Computer Vision}, 127(3):302--321,
  2019.

\bibitem{zhou2021deepvit}
Daquan Zhou, Bingyi Kang, Xiaojie Jin, Linjie Yang, Xiaochen Lian, Qibin Hou,
  and Jiashi Feng.
\newblock Deepvit: Towards deeper vision transformer.
\newblock {\em arXiv preprint arXiv:2103.11886}, 2021.

\bibitem{zhou2021refiner}
Daquan Zhou, Yujun Shi, Bingyi Kang, Weihao Yu, Zihang Jiang, Yuan Li, Xiaojie
  Jin, Qibin Hou, and Jiashi Feng.
\newblock Refiner: Refining self-attention for vision transformers.
\newblock {\em arXiv preprint arXiv:2106.03714}, 2021.

\bibitem{zoph2018learning}
Barret Zoph, Vijay Vasudevan, Jonathon Shlens, and Quoc~V Le.
\newblock Learning transferable architectures for scalable image recognition.
\newblock In {\em Proceedings of the IEEE conference on computer vision and
  pattern recognition}, pages 8697--8710, 2018.

\end{thebibliography}
}


\appendix


\end{document}